\documentclass[12pt]{book}

\usepackage{a4wide}
\usepackage{fancyhdr}
\usepackage{graphicx}
\usepackage[bookmarks]{hyperref}

\usepackage{algorithm}
\usepackage{algpseudocode}
\usepackage{booktabs}
\usepackage{caption}
\usepackage{subcaption}
\usepackage{multirow}
\usepackage{amsmath}
\usepackage{amssymb}

\usepackage{array}

\usepackage{multirow}
\usepackage{makecell}
\usepackage{pdfpages}
\usepackage{quoting}

\usepackage{booktabs}
\newcommand\norm[1]{\left\lVert#1\right\rVert}
\newcolumntype{P}[1]{>{\centering\arraybackslash}p{#1}}

\renewcommand{\chaptermark}[1]%
         {\markboth{\thechapter.\ #1}{}}
\renewcommand{\sectionmark}[1]%
         {\markright{\thesection\ #1}}
\newcommand{\LMUTitle}[9]{
  \thispagestyle{empty}
  \vspace*{\stretch{1}}
  {\parindent0cm
   \rule{\linewidth}{.7ex}}
  \begin{flushright}

    \vspace*{\stretch{1}}
    \sffamily\bfseries\Huge
    #1\\
    \vspace*{\stretch{1}}
    \sffamily\bfseries\large
    #2
    \vspace*{\stretch{1}}
  \end{flushright}
  \rule{\linewidth}{.7ex}
  \vspace*{\stretch{5}}
  \begin{center}
    \includegraphics[width=2in]{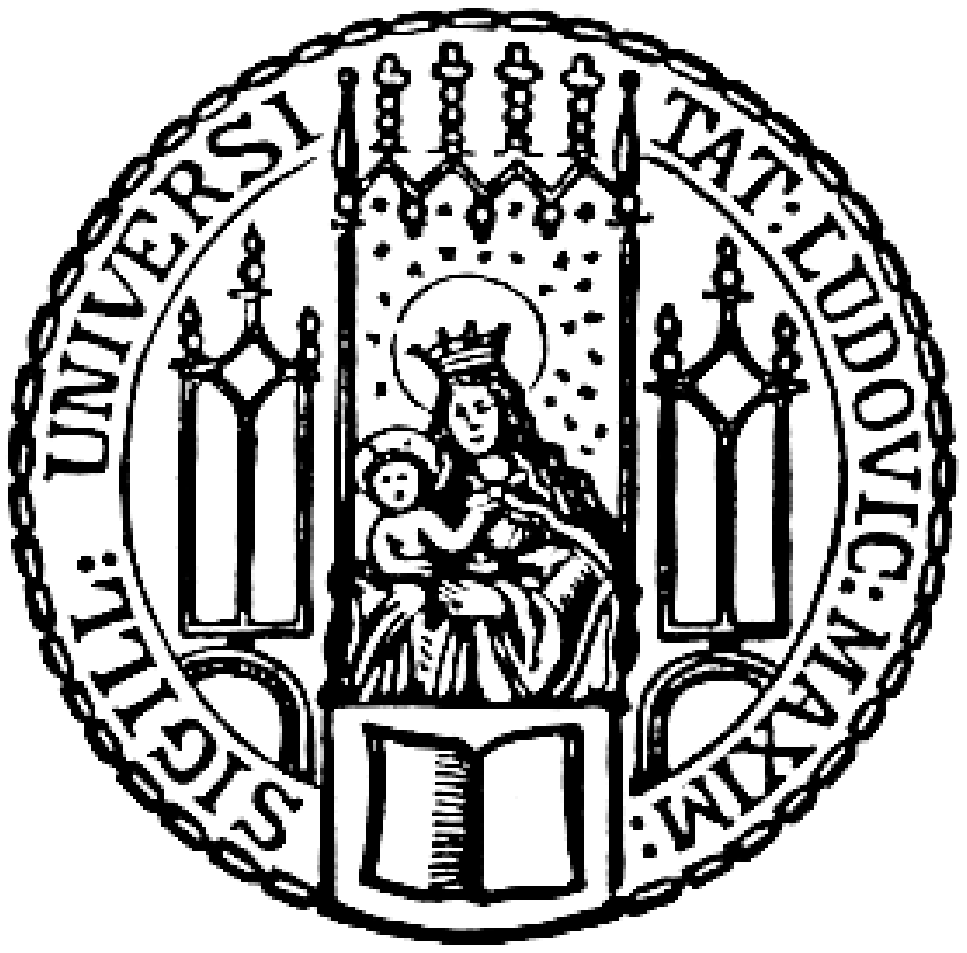}
  \end{center}
  \vspace*{\stretch{1}}
  \begin{center}\sffamily\LARGE{#5}\end{center}
  \newpage
  \thispagestyle{empty}
}

\hypersetup {
pdftitle = {Explainability and Robustness of Deep Visual Classification Models},
pdfauthor = {Jindong Gu},
pdflang = {en-US},
}

\begin{document}

  %\frontmatter

  \LMUTitle
      {Explainability and Robustness of Deep Visual Classification Models}               %\\ Titel der Arbeit
      {Jindong Gu}                       % Vor- und Nachname des Autors
      { }                             % Geburtsort des Autors
      { }                         % Name der Fakultaet
      {Munich 2022}                        % Ort und Jahr der Erstellung
      { }                            % Tag der Abgabe
      { }                          % Name des Erstgutachters
      { }                    % Name des Zweitgutachters
      { }   
      % Name des Zweitgutachters

  \markboth{Abstract}{Abstract}
  \addcontentsline{toc}{chapter}{\protect Abstract}

\section*{Abstract}
Deep learning has revolutionized AI and deep neural networks, in particular, have been hugely successful in a wide range of applications. Deep neural network architectures with different inductive biases have been proposed in different communities. In the computer vision community, Convolutional Neural Networks (CNNs), first proposed in the 1980's, have become the standard visual classification model. Recently, as alternatives to CNNs, Capsule Networks (CapsNets) and Vision Transformers (ViTs) have been proposed. CapsNets, which were inspired by the information processing of the human brain, are considered to have more inductive bias than CNNs, whereas ViTs are considered to have less inductive bias than CNNs. All three classification models have received great attention since they can serve as backbones for various downstream tasks, e.g. object detection and semantic segmentation. However, these models are far from being perfect.

As pointed out by the community, there are two weaknesses in standard Deep Neural Networks (DNNs). One of the limitations of DNNs is the lack of explainability. Even though they can achieve or surpass human expert performance in the image classification task, the DNN-based decisions are difficult to understand. In many real-world applications, however, individual decisions need to be explained. The other limitation of DNNs is adversarial vulnerability. Concretely, the small and imperceptible perturbations of inputs can mislead DNNs. The vulnerability of deep neural networks poses challenges to current visual classification models. The potential threats thereof can lead to unacceptable consequences. Besides, studying model adversarial vulnerability can lead to a better understanding of the underlying models.

Our research aims to address the two limitations of DNNs. Specifically, we focus on deep visual classification models, especially the core building parts of each classification model, e.g. dynamic routing in CapsNets and self-attention module in ViTs. 

We argue that both the lack of explainability and adversarial vulnerability can be attributed to the difference in the visual features used by visual recognition models and the human visual system to recognize objects. Namely, the visual clues used by standard CNNs are different from the ones used by our visual system. The differences make the interpretation of classifications difficult. Similarly, the differences also leave attackers the chance to manipulate decisions with quasi-imperceptible input perturbations.

We have analyzed if the brain-inspired Capsule Network (CapsNet) performs more robustly than the CNNs. Our investigation on CapsNet shows CapsNets with more inductive bias do not perform better than CNNs. The dynamic routing therein can even harm the robustness, in contrast to the common belief. Compared to CNNs and CapsNets, Vision Transformers (ViTs) are considered to have less inductive bias in their architecture. Given the patch-wise input image representation of ViT, we dissect ViT with adversarial patch attack methods. We find that vision transformers are more robust to naturally corrupted patches than CNNs, whereas they are more vulnerable to adversarial patches. Specifically, the attention module can effectively ignore naturally corrupted patches. However, when attacked by an adversary, it can be easily fooled.

Overall, our work provides a detailed analysis of CNNs, CapsNet, and ViTs in terms of explainability and robustness. The contribution of this thesis will facilitate the application of existing popular deep visual classification models and inspires the development of more intelligent classifiers in the future.

  \setcounter{page}{1}

  {\let\clearpage\relax }
  %\linespread{1.25}

\chapter{Introduction}

\section{Motivation}
Artificial intelligence changes our daily lives in many perspectives. The recent advances of artificial intelligence are mainly powered by Deep Learning method~\cite{lecun2015deep}. As a revolutionary technique, Deep Learning methods are also embraced by other disciplines, \textit{e.g.} bioscience and astronomy. As a representative model in the framework of deep learning, deep neural networks (DNNs) dominate the community due to their powerful expressiveness. However, two limitations of deep neural networks prevent their wide application in safety-critical domains, \textit{e.g.} the medical domain and autonomous driving system.

One of the limitations of deep neural networks is their lack of explainability. Even though the DNN-based intelligent system can achieve or surpass human expert performance on some tasks, it is not clear how the system reaches its decisions. For example, Deep convolutional neural networks (DCNNs) achieve start-of-the-art performance on many tasks, such as visual object recognition~\cite{Simonyan2014VeryDC,he2016deep,szegedy2016rethinking,huang2017densely}. However, since they lack transparency, they are considered as "black box" solutions. In real-world applications, however, individual decisions need to be explained to gain the trust of the users. \textit{e.g.}, autonomous driving systems should reassure passengers by giving explanations when braking the car abruptly~\cite{kim2017interpretable,kim2018textual}. Decisions made by deep models are also required to be verified in the medical domain. Mistakes of unverified models could have an unexpected impact on humans or lead to unfair decisions~\cite{liu2018delayed,hashimoto2018fairness,gu2019understanding}. Besides, AI applications must comply with related legislation, \textit{e.g.}, the right to explanation in GDPR of the European Union~\cite{selbst2018meaningful}.

The other limitation of deep neural networks is limited generalization robustness. When deep neural networks are deployed in real-world applications, the input can deviate from the training data distribution. The inference on the input with overlapped patterns~\cite{sabour2017dynamic}, affine-transformed pattern~\cite{sabour2017dynamic,gu2020improving}, and natural corruption~\cite{hendrycks2019benchmarking} can result in unexpected results. Besides the robustness to out-of-distribution data, the low robustness to artificial perturbation also raises great concern in the community. Concretely, the small and imperceptible artificial perturbations of inputs can mislead DNN-based intelligent systems. For example, given an image correctly classified by a deep convolutional neural network, a hardly human-perceptible artificial perturbation can cause the convolutional neural network to misclassify the image when added to it. The vulnerability of Deep Learning poses challenges to current intelligent systems. The adversarial images on CNNs can pose potential threats to security-sensitive CNN-based applications, \textit{e.g.}, face verification~\cite{sharif2016accessorize} and autonomous driving~\cite{eykholt2018robust}. The potential threats thereof can lead to unacceptable consequences. Besides, the existence of adversarial images demonstrates that the object recognition process in CNNs is dramatically different from that in human brains. Hence, the study of adversarial examples on deep neural networks can also lead to a better understanding of the underlying object recognition models.

\begin{figure*}[t]
    \centering
        \includegraphics[width=0.9\textwidth]{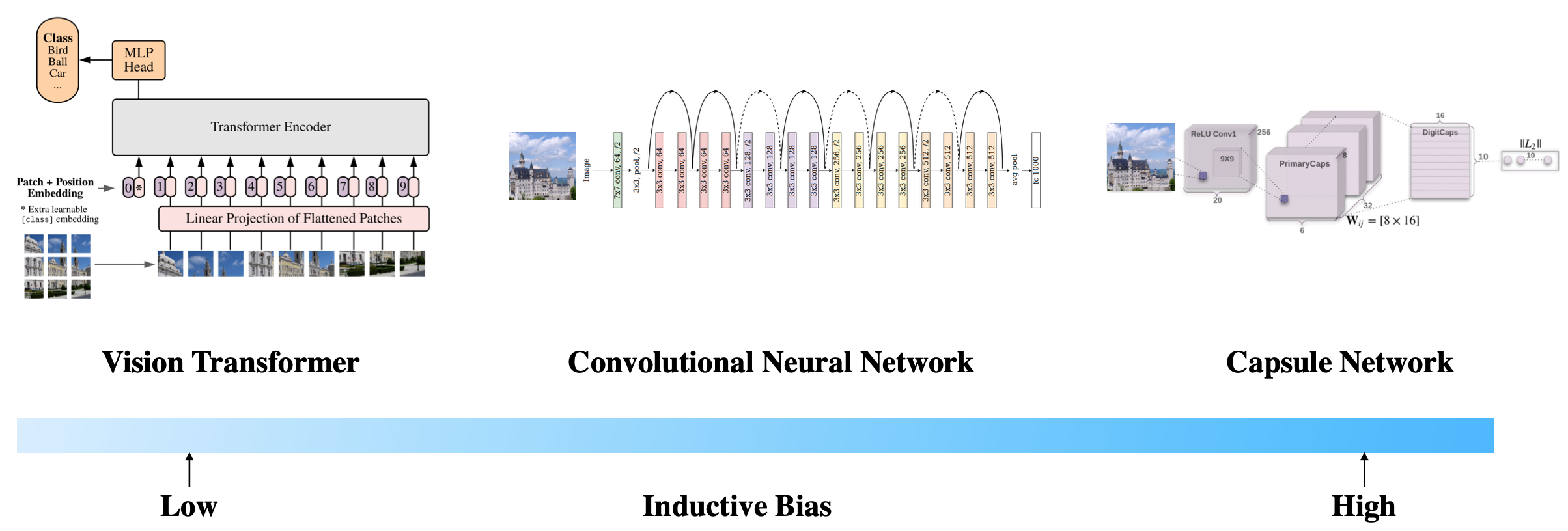}
        \caption{The overview of deep visual classification model architectures.  This figure is based on the figures in~\cite{dosovitskiy2020image,he2016deep,sabour2017dynamic}}
        \label{fig:overview}
\end{figure*}

Since~\cite{krizhevsky2012imagenet} proposed the AlexNet, deep neural networks have revolutionized the computer vision community. In the image classification task, the classification model consists of two parts, \textit{i.e.}, feature extractor and classifier. The modules that extract features from input images are also adopted as feature extractor (dubbed \textit{backbone}) in downstream tasks, \textit{e.g.}, object detection~\cite{girshick2014rich,he2017mask} and semantic segmentation~\cite{long2015fully,zhao2017pyramid,chen2017rethinking}. The improvement of the classification models often also benefits the downstream tasks due to the improved backbone. In this thesis, we focus on deep visual classification models from the perspectives of explainability and robustness.

As one of the representatives of deep visual classification models, convolutional neural networks have dominated the computer vision community in the last decade~\cite{krizhevsky2012imagenet}. However, CNNs suffer from many limitations, \textit{e.g.}, only local information aggregation at lower layers and the broken equivariance. Recently, the community has been attempting to propose new models to overcome the limitations. Two among them have received great attention from the community. The one is Capsule Networks (CapsNet) which is inspired by the information processing in the human brain~\cite{sabour2017dynamic}. Compared to CNNs, CapsNet is more inductively-biased where the partial information processing in the human brain is integrated into the model, \textit{e.g.}, the transformation process. The other is Vision Transformer(ViT)~\cite{dosovitskiy2020image}. Given the success of Transformer in natural language processing (NLP), the work~\cite{dosovitskiy2020image} generalizes Transformer architectures to image classification task by representing the input image as a sequence of image patches. Compared to CNNs, ViTs are less inductive-biased where information aggregation is also possible at lower layers. Convolutional Neural Networks, Capsule Networks, and Vision Transformers raise great attention in the community. Hence, in this work, we mainly focus on the three deep visual classification models.

In the rest of this chapter, we first introduce background knowledge about CNNs, CapsNets, and ViTs in Section~\ref{sec:background}. Then, in Section~\ref{sec:explain}, we present a summary of the explainability of deep visual classifications and describe our contributions to the explainability of deep visual classification models. Last, in Section~\ref{sec:robust}, we show the categorization of the robustness of deep visual classifications and describe our contributions to the robustness of deep visual classification models.

\textbf{Contributions.} In this dissertation, our contributions can be summarized from two perspectives. From the perspective of explainability, we first present a novel method, called CLRP, to explain CNN-based image classifications in Chapter 2. Then, in Chapter 3, we present our interpretable capsule networks whose predictions can be explained with built-in modules. Last, we show our understanding of ViT-based image classifications in Chapter 7. From the perspective of robustness, our contributions mainly focus on the role the model architecture plays in terms of both natural robustness and adversarial robustness. We present our findings and improvements of Capsule Networks' natural robustness to non-additive perturbation in Chapters 4 and 5, and further propose our adversary Vote Attack method to show the vulnerability of CapsNets in Chapter 6. Besides, we introduce our understanding of the robustness of ViT-based classifications to patch-wise perturbations in Chapter 7.

  \section{Background Knowledge}
\label{sec:background}

\subsection{Convolutional Neural Networks}
To recognize the patterns of the images, many operations have been proposed, e.g., Scale-Invariant Feature Transform (SIFT)~\cite{lowe1999object}, Histogram of Oriented Gradients(HOG)~\cite{dalal2005histograms}, and Convolution. Especially, the convolutional operation dominates the community in the last decade as an image feature extraction operation.

Formally, convolution is a mathematical operation on two functions that produces a third function that expresses how the shape of one is modified by the other. In the domain of computer vision, the discrete variant of convolution is adopted since the images are saved as discrete signals. Concretely, given an image $\boldsymbol{X} \in \mathbb{R}^{(C \times H \times W)}$ and a convolution kernel $\boldsymbol{k} \in \mathbb{R}^{(C\times P\times Q)}$, the feature map $\boldsymbol{H} \in \mathbb{R}^{(H' \times W')}$ extracted by the convolution kernel is computed as
\begin{equation}
\boldsymbol{H}_{(i,\;\;  j)}=\sum^C_{c=1} \sum^P_{p=1} \sum^Q_{q=1} \boldsymbol{X}_{(c,\;\; i+p-1, \;\; j+q-1)} \; \boldsymbol{k}_{(c,\;\; p,\;\;  q)},
\label{equ:cov}
\end{equation}
where $(i,j)$ is the index of elements in the feature map $\boldsymbol{H}$, $C$ is the number of channels of input images and $(P, Q)$ are the size of the feature map. A single kernel corresponds to a single feature map. Multiple kernels are often applied to extract multiple feature maps. 

Besides, the pooling (subsampling) operation is applied to the feature maps extracted by convolution operation to aggregate the visual information. In the pooling operation, the mean operation or the max operation is often applied. The pooling operation with size $(s, s)$ can be expressed as 
\begin{equation}
\boldsymbol{H}'_{(i,\;\;  j)}= \max^P_{p=1} \boldsymbol{H}_{(i,\;\;  j)}.
\label{equ:pool}
\end{equation}
Convolution can be further applied to the pooled feature maps. The convolutional and pooling operations are applied alternatively on the image to obtain the final feature maps.

The features $\boldsymbol{H}^L_{(i,\;\;  j)}$ extracted by a list of convolutional operations and pooling operations are taken as the final image representation. A single or multiple fully connected layers (i.e. a MLP module) is used as classifier that maps the features into the ground-truth class.
\begin{equation}
\boldsymbol{Z} = MLP(\boldsymbol{H}^L_{(i,\;\;  j)})
\label{equ:pool}
\end{equation}
The output probabilities can be obtained by applying softmax function on the logits $\boldsymbol{Z}$. The predicted class is defined as $argmax(\boldsymbol{Z}_i)$.

\begin{figure*}[!h]
    \centering
        \includegraphics[width=1.0\textwidth]{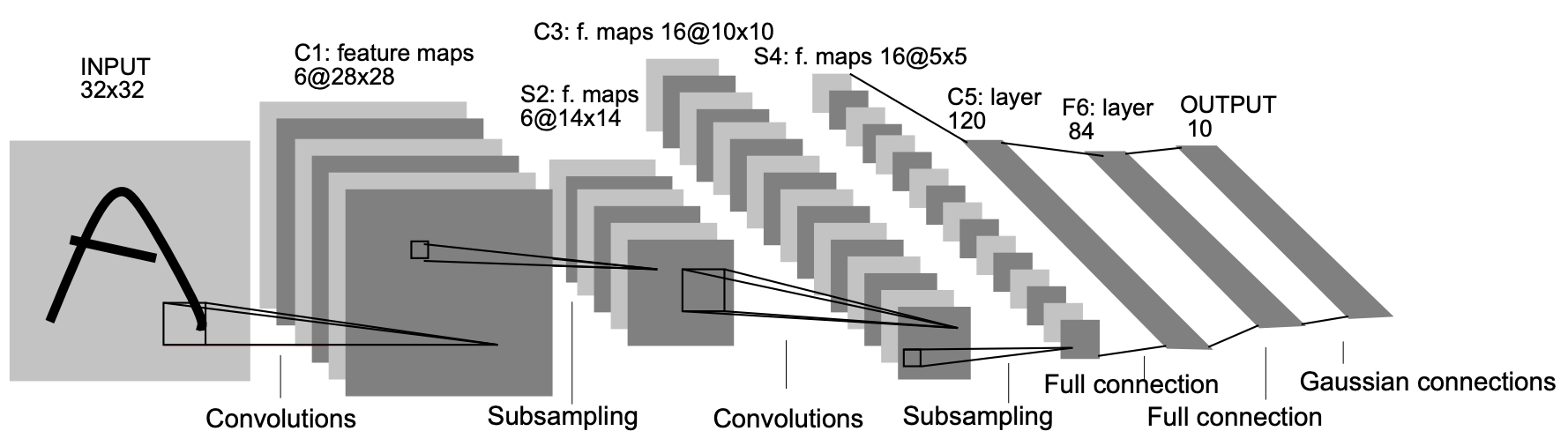}
        \caption{The overview of LeNet-5 architecture~\cite{lecun1998gradient}.}
        \label{fig:lenet}
\end{figure*}

The work~\cite{lecun1998gradient} proposes Convolution Neural Network (CNN) in the end-to-end learning framework to recognize hand-written digits. Therein, LeNet-5 is the classic instance of convolution neural networks, which is visualized in Fig.~\ref{fig:lenet}. The proposed LeNet-5 starts with two convolutional layers, and each is followed by a pooling layer. Then, a three-layer MLP module maps the feature to the logits.

\begin{figure*}[!h]
    \centering
        \includegraphics[width=1.0\textwidth]{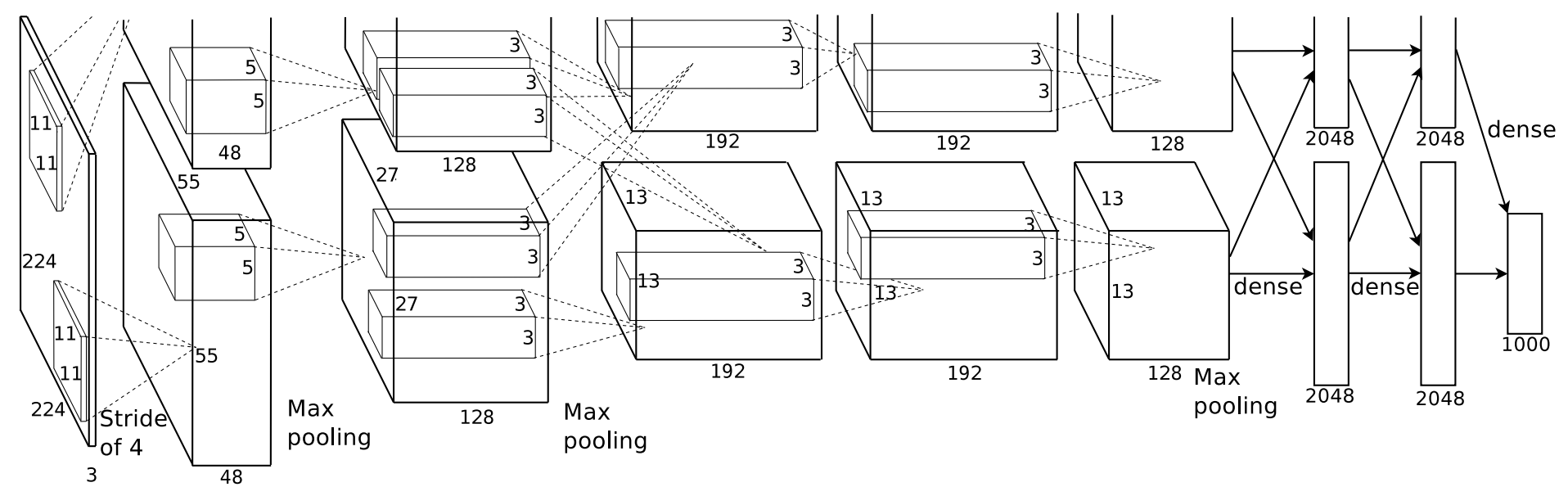}
        \caption{The overview of AlexNet architecture~\cite{krizhevsky2012imagenet}.}
        \label{fig:alexnet}
\end{figure*}

Given the limited computational resource, the architecture and the corresponding training strategy proposed in~\cite{lecun1998gradient} does not scale well to the large-scale dataset. With the advance of the computational power, the work~\cite{krizhevsky2012imagenet} proposes AlexNet, which achieves impressive accuracy on ImageNet-1k dataset. AlexNet consists of five convolutional layers, some of which are followed by max-pooling layers,
and three fully-connected layers with a final 1000-way softmax. In terms of model architecture, AlexNet is deeper and wider than LeNet-5. From the perspective training strategy, to make AlexNet work well, the work~\cite{krizhevsky2012imagenet} proposes non-saturating neurons, i.e., Rectified Linear Units (ReLUs) to activate the neurons and employs dropout method to regularize the training process. Especially, they propose a GPU-specific implementation of GPU operation to make the training process feasible.

\begin{figure*}[t]
    \centering
        \includegraphics[width=0.5\textwidth]{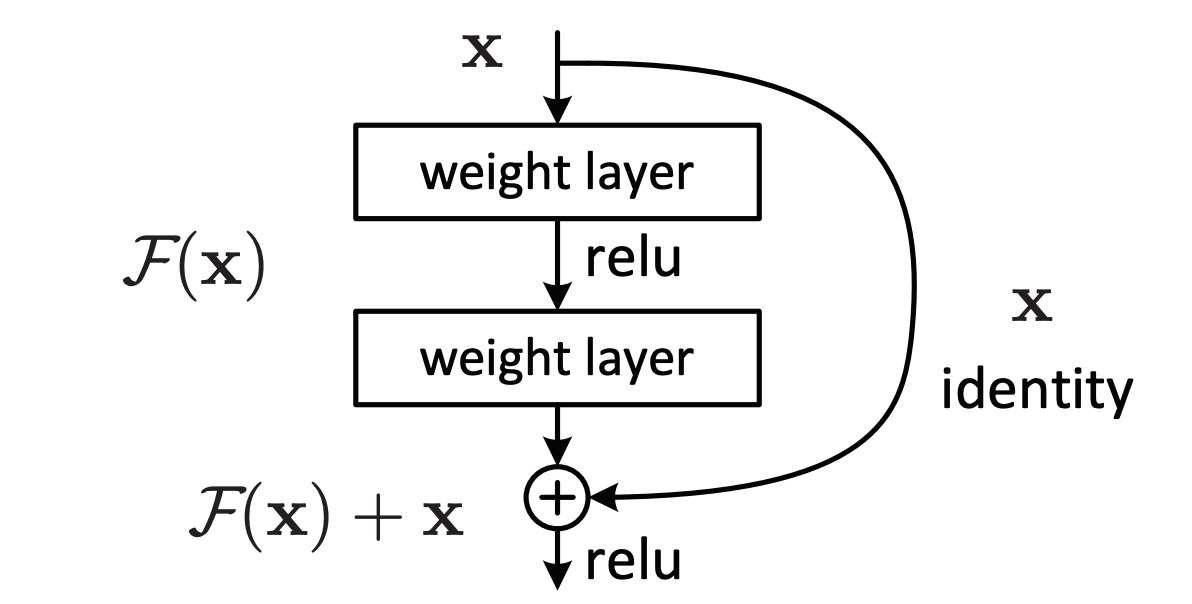}
        \caption{The overview of Residual block with skip connection~\cite{he2016deep}.}
        \label{fig:skip}
\end{figure*}

\begin{figure*}[t]
    \centering
        \includegraphics[width=1.0\textwidth]{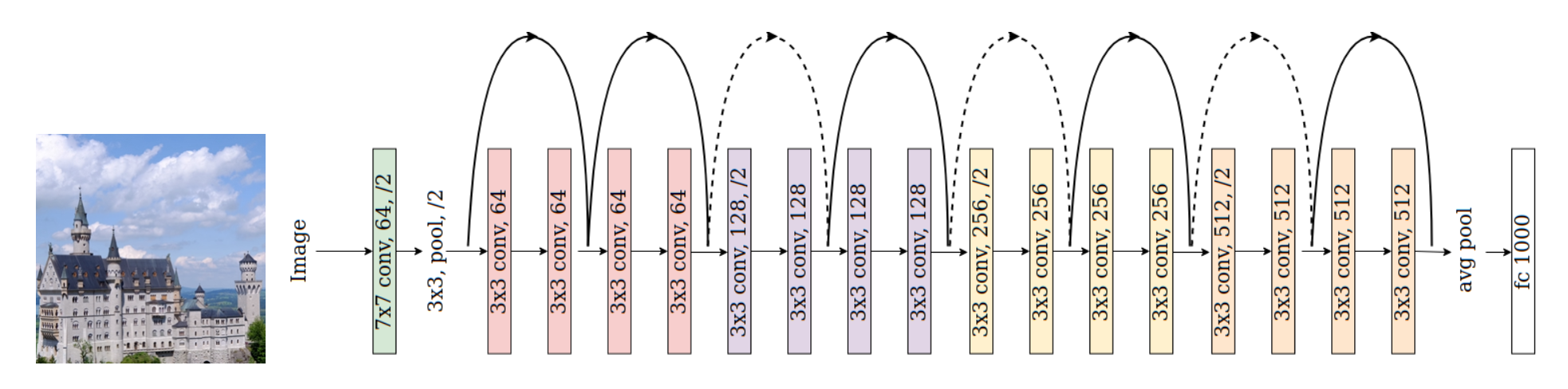}
        \caption{The overview of ResNet architecture~\cite{he2016deep}.}
        \label{fig:resnet}
\end{figure*}

One intuitive way to improve AlexNet is to build deeper layers. However, the AlexNet with deeper layers does not converge well during training due to the gradient vanishing problem. Namely, the gradients become zeros or close to zeros when propagating from the output layer to low layers. Due to the gradient vanishing problem, the parameter update of low layers is challenging. To overcome the challenges, the work~\cite{he2016deep} proposes skip-connection, which can propagate the gradients from deep layers to low layers directly by skipping some intermediate layers.

The block with such a skip connection is called residual block. A popular residual block is shown in Fig.~\ref{fig:skip}. As an instance, the work~\cite{he2016deep} proposes ResNet which consists of a list of residual blocks. When equipped with skip connections, ResNets with even more than 100 layers can converge well. ResNets still dominate the computer vision community. We show the ResNet18 in Fig.~\ref{fig:resnet} as an example where 18 layers are built into the ResNet to extract features.

\paragraph{Convolutional Network Follow-Ups:} The CNN-based deep visual classifier has already surpassed human-level performance in the image classification task~\cite{khan2020survey}. In the last years, the architectures of convolutional neural networks have still been improved from different perspectives. On the one hand, the more advanced architectures have been proposed to further push the state-of-the-art performance~\cite{szegedy2016rethinking,huang2017densely,dai2017deformable}. On the other hand, the efficiency of architecture has received great attention since real-world CNN-based applications often require less memory consumption and computational cost. The efficiency of architecture has been addressed from different perspective, e.g., light-weight architecture design~\cite{howard2017mobilenets,zhang2018shufflenet}, architecture pruning~\cite{lecun1990optimal,hassibi1993second,han2015learning,Molchanov2017PruningCN}, and distilling knowledge from large architectures to small architectures~\cite{hinton2015distilling,Romero2015FitNetsHF,gu2020search,gu2021simple}. More recently, many researchers focus on neural architecture search where the architectures are searched automatically from a predefined search space~\cite{Zoph2017NeuralAS,Liu2018HierarchicalRF,Liu2019DARTSDA}. The found architecture can surpass the manually designed ones.

\begin{figure*}[t]
    \centering
        \includegraphics[width=1.01\textwidth]{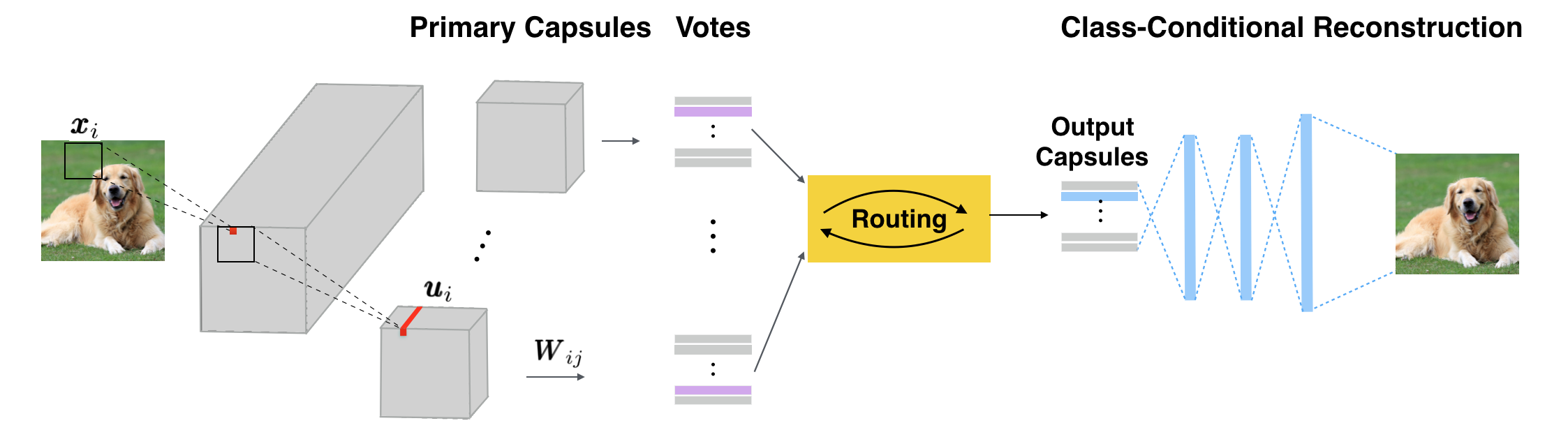}
        \caption{The overview of CapsNet architectures.  The CapsNet architecture consists of four components, such as primary capsule extraction, voting, routing, and class-conditional reconstruction. The primary capsule extraction module first maps the raw input features to low-level capsules. The voting process transforms low-level capsules to make votes with a transformation matrix. Then, the routing module identifies the weight of each vote and computes the final high-level capsules. In the last part, the reconstruction subnetwork reconstructs input images from capsules to regularize the learning process.}
        \label{fig:overview_capsnet}
\end{figure*}

\subsection{Capsule Networks}
Inspired by the information process in the human brain, Hinton proposes Capsule Networks (CapsNet)~\cite{sabour2017dynamic}. Different from CNNs, CapsNets represent a visual entity with a vector instead of a single scale value, called Capsule. 
CapsNets \cite{sabour2017dynamic} encode visual entities with capsules. Each capsule is represented by an activity vector (e.g., the activation of a group of neurons), and elements of each vector encode the properties of the corresponding entity. The length of the activation vector indicates the confidence of the entity's existence. The output classes are represented as high-level capsules.

The most popular version of Capsule Networks is Dynamic Routing Capsule Networks (DR-CaosNet). We introduce the architecture details of DR-CapsNet as follows. As shown in Fig.~\ref{fig:overview_capsnet}, CapsNet starts with one (or more) convolutional layer(s) that convert the raw pixel intensities $\boldsymbol{X}$ into low-level visual entities $\boldsymbol{u}_i$. Concretely, CapsNet extracts feature maps of shape $(C', H', W')$ from input image $\boldsymbol{X} \in \mathbb{R}^{(C \times H \times W)}$ with two standard convolutional layers where $C'$, $H'$, $W'$ are the number of channels, the height, and the width of the feature maps, respectively. The extracted feature maps are reformulated as primary capsules $(C'/D_{in}, H', W', D_{in})$ where $D_{in}$ is the dimensions of the primary capsules. There are $N = C'/D_{in} * H' * W'$ primary capsules all together. Each capsule $\boldsymbol{u}_i$, a $D_{in}$-dimensional vector, consists of $D_{in}$ units across $D_{in}$ feature maps at the same location. For example, the red bar marked with $\boldsymbol{u}_i$ in Fig.~\ref{fig:overview_capsnet} is a low-level capsule.

In the voting process, each primary capsule is transformed to make a vote with a transformation matrix $\boldsymbol{W}_{ij} \in \mathbb{R}^{(D_{in} \times N*D_{out})}$ in, where $N$ is the number of output classes and $D_{out}$ is the dimensions of output capsules. The vote from the $i$-th low-level capsules to the $j$-th high-level capsules is
\begin{equation}
\boldsymbol{\hat{u}}_{j|i}=\boldsymbol{W}_{ij} \boldsymbol{u}_i.
\label{equ:trans}
\end{equation}

Then, a routing module is applied to identify weight for each vote. Given all $N$ votes $\boldsymbol{\hat{u}}_{j|i}$ of the $L$-th layer with $N$ capsules, $M$ high-level capsule $\boldsymbol{s}_j$ of the $(L+1)$-th layer with $M$ capsules, the routing process is
\begin{equation}
\boldsymbol{s}_j=\sum^N_i c_{ij} \boldsymbol{\hat{u}}_{j|i}
\label{equ:routing}
\end{equation}
where $c_{ij}$ is a coupling coefficient that models the degree with which $\boldsymbol{\hat{u}}_{j|i}$ is able to predict $\boldsymbol{s}_j$. The capsule $\boldsymbol{s}_j$ is shrunk to a length in [0, 1) by a non-linear squashing function $g(\cdot)$, which is defined as
\begin{equation}
\boldsymbol{v}_j= g(\boldsymbol{s}_j) = \frac{\norm{\boldsymbol{s}_j}^2}{1+\norm{\boldsymbol{s}_j}^2} \frac{\boldsymbol{s}_j}{\norm{\boldsymbol{s}_j}}.
\label{equ:squa}
\end{equation}
By doing the squashing operation, the length of the vector is mapped to [0, 1) that represents the confidence of the high-level entity's existence. In DR-CapsNet, the high-level capsules correspond to output classes, and its length means the output probability.

Note that the coupling coefficients $\{c_{ij}\}$ in Equation~\ref{equ:routing} are computed by an iterative routing procedure. They are updated so that high agreement ($a_{ij} =  \boldsymbol{v}^T_j \boldsymbol{\hat{u}}_{j|i}$) corresponds to a high value of $c_{ij}$. 
\begin{equation}
c_{ij}= \frac{\exp(b_{ij})}{\sum_k \exp(b_{ik})}
\label{equ:coup}
\end{equation}
where initial logits $b_{ik}$ are the log prior probabilities and updated with $b_{ik} = b_{ik} + a_{ij} $ in each routing iteration. The coupling coefficients between a $i$-th capsule of the $L$-th layer and all capsules of the $(L+1)$-th layer sum to  1, i.e., $\sum_{j=1}^M c_{ij} =1$. The steps in Equations \ref{equ:trans}, \ref{equ:routing}, \ref{equ:squa}, and \ref{equ:coup} are repeated $K$ times in the routing process, where $\boldsymbol{s}_j$ and $c_{ij}$ depend on each other.

The length of the final output capsule $\boldsymbol{v}_j$ corresponds to the output probability of the $j$-th class. Different from CNNs where cross-entropy loss is often applied to compute classification loss. In DR-CapsNet, the margin loss function is applied to compute the classification loss
\begin{equation}
\begin{split}
L_k = & T_k  \max(0, m^+ - \norm{\mathbf{v}_k})^2 \\
& + \lambda (1 -T_k) \max(0, \norm{\mathbf{v}_k} -  m^-)^2
\end{split}
\label{equ:marginloss}
\end{equation}
where $T_k = 1$ if the object of the $k$-th class is present in the input. As in~\cite{sabour2017dynamic}, the hyper-parameters are often empirically set as $m^+ = 0.9$, $m^- = 0.1$ and $\lambda=0.5$.

A reconstruction sub-network reconstructs the input image from all $N$ output capsules with a masking mechanism. The ones corresponding to the non-ground-truth classes are masked with zeros before being transferred to the reconstruction sub-network. Due to the masking mechanism, only the capsule of the ground-truth class is visible for the reconstruction. Hence, the reconstruction process is called class-conditional reconstruction. The reconstruction loss is computed as a regularization term in the loss function.

\paragraph{Capsule Network Follow-Ups:} Many routing mechanisms have been proposed to improve the performance of CapsNet, such as Expectation-Maximization Routing~\cite{hinton2018matrix}, Self-Routing~\cite{hahn2019self}, Variational Bayes Routing~\cite{ribeiro2020capsule}, Straight-Through Attentive Routing~\cite{ahmed2019star}, and Inverted Dot-Product Attention routing~\cite{Tsai2020Capsules}. An alternative to the routing mechanism to aggregate information is proposed in work~\cite{gu2020interpretable} where they replace the dynamic routing with a multi-head attention-based graph pooling approach. To reduce the parameters of CapsNet, a matrix or a tensor is used to represent an entity instead of a vector~\cite{hinton2018matrix,rajasegaran2019deepcaps}. The size of the learnable transformation matrix can also be reduced by the matrix/tensor representations. Besides, the work~\cite{gu2020improving} proposes to share a transformation matrix to reduce the network parameters. Another way to improve CapsNet is to integrate advanced modules of ConvNet into CapsNet, e.g., skip connections~\cite{he2016deep,rajasegaran2019deepcaps} and dense connections~\cite{huang2017densely,phaye2018multi}.

\begin{figure*}[t]
    \centering
        \includegraphics[width=0.8\textwidth]{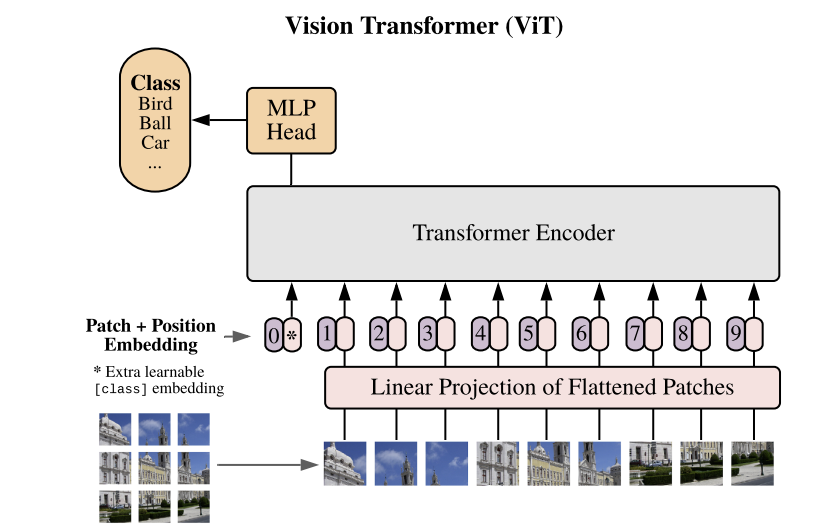}
        \caption{The overview of Vision Transformer Architectures. The figure is taken from~\cite{dosovitskiy2020image}.}
        \label{fig:overview_vit}
\end{figure*}

\begin{figure*}[t]
    \centering
        \includegraphics[width=0.8\textwidth]{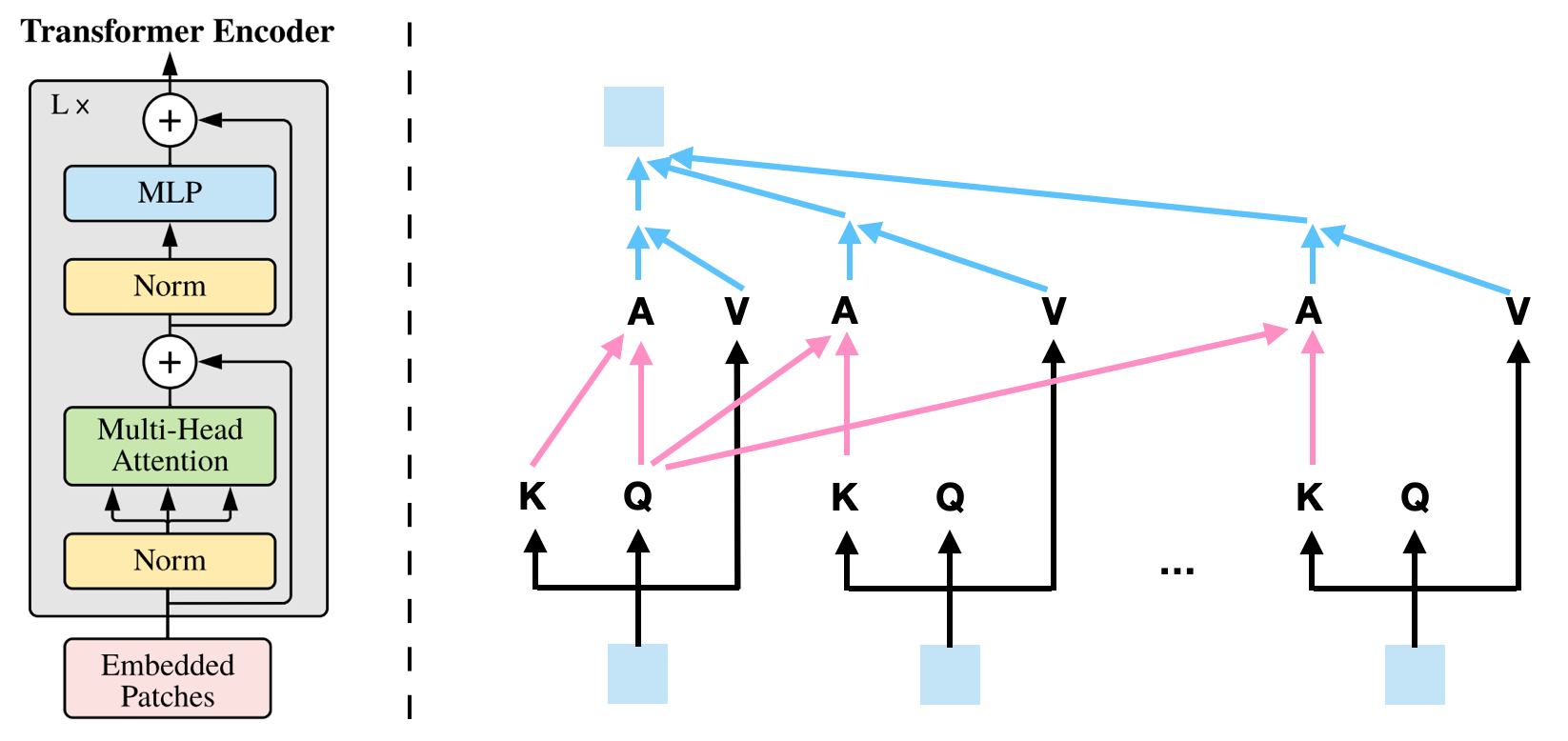}
        \caption{The overview of Transformer Encoder.}
        \label{fig:overview_trans_encoder}
\end{figure*}

\subsection{Vision Transformers}
Transformers with self-attention-based architectures have become the model of choice in natural language processing (NLP)~\cite{vaswani2017attention}. Inspired by the success of Transformers in NLP community, the work~\cite{dosovitskiy2020image} proposes Vision Transformer(ViT) where they replace
the convolutions entirely with self-attention layers and achieve remarkable performance in the image classification task. As a promising alternative to CNNs, Vision Transformer raises the great attention of our community.

Different from CNNs, ViT represents an input image as a sequence of image patches. Then, the list of self-attention modules are applied to the sequence of image patches sequentially. We now introduce the details of the primary Vision Transformer architecture in~\cite{dosovitskiy2020image}. As shown in Fig.~\ref{fig:overview_vit}, the input image $\boldsymbol{X} \in \mathbb{R}^{(C \times H \times W)}$ is split into image patches $\{\boldsymbol{x}_i\in \mathbb{R}^{P\times P\times C}\vert i \in (1,2,3, ..., {H/P}\times{W/P}) \}$ where $P$ is the patch size. The embedding of each patch is extracted from the raw image patch with linear projection parameters $\boldsymbol{W}^0 \in \mathbb{R}^{(HW/P^2\times D_p)}$. Before the application of self-attention module, the position information of image patches is also integrated into the patch embedding. The embedding of the patch $\boldsymbol{x}_i$ is described as  
\begin{equation}
\boldsymbol{E}^0_i= \boldsymbol{x}_i \cdot \boldsymbol{W}^0 + \boldsymbol{PE}_i,
\label{equ:trans}
\end{equation}
where $\boldsymbol{PE}_i$ is the position embedding of the image patch $\{\boldsymbol{x}_i$, which encodes the patch position information in the input images. The position embedding $\boldsymbol{PE}_i$ could be manually designed or learnable. In ViT, the learnable version is adopted.

A learnable class-token embedding $\boldsymbol{E}^0_0$ is added into the list of patch embeddings. The class embedding in the last layer is taken as the image embedding for classification. We now introduce the transformer encoder where the list of blocks is applied to transform the input embeddings. As shown in Fig.~\ref{fig:overview_trans_encoder}, each block consists of two main modules, namely, a multi-head self-attention module to model the inter-patch relationship and an MLP module to project each patch respectively.

When the self-attention module with a single head in $l+1$-th layer is applied to input patches $\{\boldsymbol{E}^l_i\in \mathbb{R}^{D_p}\vert i \in (0,1,2,3, ..., {H/P}\times{W/P})\}$ in the $l$-th layer, the output embedding of the patch $\boldsymbol{E}^l_i$ is  
\begin{equation}
\begin{split}
\boldsymbol{K}^{l+1}_i &= \boldsymbol{W}^{l+1}_k \cdot \boldsymbol{E}^l_i,  \\ \boldsymbol{Q}^{l+1}_i &= \boldsymbol{W}^{l+1}_q \cdot \boldsymbol{E}^l_i, \\
\boldsymbol{V}^{l+1}_i &= \boldsymbol{W}^{l+1}_v \cdot \boldsymbol{E}^l_i, \\
\boldsymbol{A}^{l+1}_i = Softmax(\boldsymbol{Q}^{l+1}_i \cdot  \boldsymbol{K}^{l+1}_0,& \quad  \boldsymbol{Q}^{l+1}_i \cdot \boldsymbol{K}^{l+1}_1, \quad  ..., \quad\boldsymbol{Q}^{l+1}_i \cdot \boldsymbol{K}^{l+1}_{{H/P}\times{W/P}+1},), \\
\boldsymbol{E}^{l+1}_i &= \sum^{{H/P}\times{W/P}+1}_{j=1} \boldsymbol{A}^{l+1}_{ij} \cdot \boldsymbol{V}_{j}.
\end{split}
\label{equ:self_attention}
\end{equation}
In this equation, the key, query, and value of patch embedding is computed first. The attention of $\boldsymbol{E}^{l+1}_i$ to all patches in $l$-th layer is obtained with the query of $i$-th patch and all keys. The output embedding $\boldsymbol{E}^{l+1}_i$ is the weighted sum of all values of patches. The output embeddings of different heads are concatenated as the final embedding. Then, an MLP module with two MLP layers is applied to project the final embedding of each patch into a new feature space. The final embedding of the class-token patch is taken as the image representation to classify the image. A linear classifier maps the features to output space.

\paragraph{Vision Transformer Follow-Ups:} Since the ViT was proposed, many new vision transformer architectures have been proposed~\cite{touvron2021training,han2021transformer,liu2021swin}. A hybrid architecture that consists of both convolutional layers and self-attention blocks has also been explored~\cite{graham2021levit,xiao2021early}. Besides, the pure patch-based architecture without attention mechanism has also been proposed~\cite{tolstikhin2021mlp}. By the time this thesis is written, the arm-race between ResNet and Vision Transformers is still going on~\cite{liu2022convnet}. Recently, many researchers employ the Transformer architecture as a uniform architecture that model both images and texts~\cite{radford2021learning,wang2021vlmo}.

%%%%%%%%%%%%%%%%%%%%%%%%%%%%%%
%%  Einbinden einer Grafik  %%
%%%%%%%%%%%%%%%%%%%%%%%%%%%%%%

  \section{Explanability of Deep Visual Classifications}
\label{sec:explain}

\subsection{Introduction}
Deep Neural Networks (DNNs) have shown impressive performance in high-dimensional input data. Especially, the performance of DNNs can even surpass human-level performance in the image classification task. The traditional machine learning methods classify images with hand-crafted images, while DNNs make predictions based on the features learned automatically from data with an optimization algorithm. Hence, it is challenging to understand the classification decisions made by DNNs. In recent years, many directions have been explored to explain individual image classifications. We summarize and roughly categorize them in Table \ref{tab:summary}. We introduce each approach as follows.

\begin{table}[]
\centering
\begin{tabular}{ | m{5.3cm} | m{10cm} | }
\hline
\multicolumn{1}{|c|}{\textbf{Approach}} &  \multicolumn{1}{|c|}{ \textbf{Description}} \\
\hline
Saliency Maps & Identifying the relevance of each input pixel to the output class \cite{Simonyan2013DeepIC,bach2015pixel,selvaraju2017grad,shrikumar2017learning,sundararajan2017axiomatic,smilkov2017smoothgrad,Gu2018UnderstandingID,srinivas2019full,dabkowski2017real,schwab2019cxplain,ribeiro2016should,zintgraf2017visualizing,gu2022verification,fong2017interpretable,gu2019contextual}. \\
\hline
Counterfactual Explanation & Identifies how the given input could change such that the classifier would output a different specified class~\cite{Chang2018ExplainingIC,Goyal2019CounterfactualVE}. \\
\hline
Explanatory Sentences & Generating natural language sentences that describe the class-discriminative pixels~\cite{hendricks2016generating,hendricks2018grounding}.  \\
\hline
Supporting Training Images & Identifying training images most responsible for a given prediction~\cite{koh2017understanding}. \\
\hline
Built-in Explanation & Generating Explanations with built-in modules (in explainable classifier) for a given prediction~\cite{koh2017understanding}.  \\
\hline
Disentangled Representations & Identifying the human-interpretable properties of the recognized object in the input image~\cite{sarhan2019learning,sabour2017dynamic,jung2020icaps,zhu2021and}.  \\
\hline
\end{tabular}
\caption{Summarization of different approaches for explaining image classifications.} 
\label{tab:summary}
\end{table}

Saliency Maps, as intuitive explanations, have received great attention in the community. The saliency map is a heat map, each element of which indicates the importance of the pixel in the corresponding position. The saliency map is expected to have recognizable patterns like the objects in the input image. The primary work~\cite{Simonyan2013DeepIC} takes the vanilla gradient of the loss with respect to the input as the saliency map. However, the gradients are noisy and the pattern therein is barely recognizable. To improve the saliency map, many methods have been proposed~\cite{Simonyan2013DeepIC,springenberg2014striving,bach2015pixel,selvaraju2017grad,shrikumar2017learning,sundararajan2017axiomatic,smilkov2017smoothgrad,Gu2018UnderstandingID,srinivas2019full,gu2019saliency,gu2019semantics}. The primary method and the improved variants are model-aware, which leverage the parameters and the gradients of neural networks to compute saliency maps. Besides the model-aware methods, the model-agnostic saliency methods are also preferred in many scenarios. For example, they are able to explain any classifiers; the explanations produced from two or more different types of models are comparable; an ensemble model can be explained without requiring knowledge of model components. There are two types of model-agnostic saliency methods. The one is to build an explanation generation model, e.g. a neural network with U-net architecture~\cite{ronneberger2015u,dabkowski2017real,schwab2019cxplain}. The other is to approximate the local decision boundary of the underlying model with an explainable model, e.g., linear classifier~\cite{ribeiro2016should}. The explanation generated from the explainable surrogate model can be used to explain individual decisions.

Counterfactual Explanation describes what changes to the situation would have
resulted in arriving at the alternative decision. In the case of image classification, Counterfactual Explanation is the counterfactual image, which indicates that the output will become the target class if the input image is replaced with the counterfactual image. The work~\cite{Chang2018ExplainingIC} creates a counterfactual image with a conditional generative model, which generates part of the pre-defined image region conditional on the rest of the image. The desired property of the generated image is to most change the classifier’s decision. Another work~\cite{Goyal2019CounterfactualVE} formulates the generation of the counterfactual image as an image editing problem. Their method performs well even in the fine-grained classifications.

Natural language, as a natural interface, has also been explored to explain the visual classifications. The works~\cite{hendricks2016generating,hendricks2018grounding} build modules to generate natural language sentences to explain the decisions where the sentences describe the class-discriminative features. The explanatory sentences are different from the caption/description generated by multi-model models. The contemporary vision-language models describe image content but fail to tell class-discriminative features which justify visual predictions.

Another way to explain visual classifications is to identify the training points most responsible for a given prediction. To trace a model’s prediction back to its training data, the work~\cite{koh2017understanding} leverages influence functions, i.e., a classic technique from robust statistics. Given a classification, they can be the most responsible training image that supports the predictions. The created explanation can tell where the local decision boundary of the model came from at a specific data point.

The approaches introduced above are post-hoc. Namely, the explanations are created for off-shelf models without intervening in their training process. An alternative to post-hoc explanation methods is to integrate dedicated modules into the model to be trained, e.g. attention mechanism~\cite{gu2020interpretable}, explanation module~\cite{dabkowski2017real} and prototype module~\cite{chen2019looks}. In the inference stage, the modules can be used to create explanations directly. The created explanations are dubbed built-in explanations, which are more efficient and easy to create.

The image representations learned by DNNs are often distributed, which makes the classification decision less explanation. It is difficult to interpretable the decision process inside the model. One way to mitigate this problem is to constrain the model to learn disentangled representations where each element of representation corresponds to a human-understandable concept~\cite{sarhan2019learning,sabour2017dynamic,jung2020icaps,gu2019neural,zhu2021and}.

%\textbf{Contributions.} In this subsection, we have introduced the existing popular methods applied to explain individual classification decisions. In the rest of this section, we present our contributions towards understanding the classifications. Specifically, in section~\ref{subsec:cnn_exp}, we present a novel method, called CLRP, to explain CNN-based image classifications. We elaborate our CLRP explanation approcah in Chapter 2. Next, in section~\ref{subsec:caps_exp}, we describe our interpretable capsule networks whose predictions can be explained with built-in modules, which is elaborated in Chapter 3. Last, we introduce our understanding of ViT-based image classifications in section~\ref{subsec:vit_exp} and elaborate more details in Chapter 7.

In this subsection, we have introduced the popular methods applied to explain individual classification decisions. In the rest of this section, we present our contributions towards
understanding the classifications. Specifically, we briefly introduce our works on the topic
of explaining classification decisions made by Convolutional Networks, Capsule Networks,
and Vision Transformers.

\subsection{Explainability of Convolutional Neural Network-based Classification}
\label{subsec:cnn_exp}
A large number of saliency methods have been proposed to better understand individual decisions of deep convolutional neural networks. As one of the representatives, the Layer-wise Relevance Propagation (LRP) approach is able to create pixel-wise explanatory saliency maps. LRP method has also been widely applied to many tasks in different domains, e.g., in medical domain~\cite{yang2018explaining} and in NLP~\cite{arras2017explaining}.

The explanations generated by LRP are known to be pixel-wise and instance-specific. However, the discriminativeness of the explanations has not been evaluated yet. Ideally, the visualized objects in the explanation should correspond to the class that the class-specific neuron represents. Namely, the explanations should be class-discriminative.

\begin{figure}[t]
    \centering
    \includegraphics[scale=0.28]{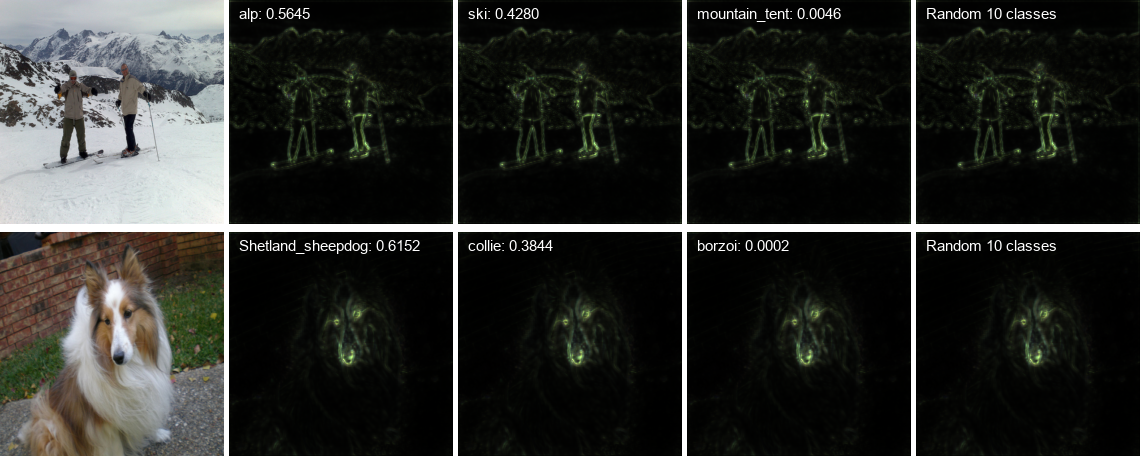}
    \caption{The explanations generated by LRP on VGG16 Network. The images from validation datasets of ImageNet are classified using the off-the-shelf models pre-trained on the ImageNet. The classifications of the images are explained by the LRP approach. For each image, we generate four explanations that correspond to the top-3 predicted classes and a randomly chosen multiple-classes. The explanations are not class-discriminative.}
\label{fig:lrp_eval}
\end{figure}

Our work~\cite{Gu2018UnderstandingID} evaluates the discriminativeness of the explanations generated by LRP. Concretely, we evaluate the explanations generated by LRP on the off-the-shelf models, e.g., VGG16 \cite{Simonyan2014VeryDC} pre-trained on the ImageNet dataset \cite{deng2009imagenet}. The results are shown in Fig. \ref{fig:lrp_eval}. For each test image, we create four saliency maps as explanations. The first three explanation maps are generated for top-3 predictions, respectively. The fourth one is created for randomly chosen 10 classes from the top-100 predicted classes (which ensure that the score to be propagated is positive). The white text in each explanation map indicates the class the output neuron represents and the corresponding classification probability. The generated explanations are instance-specific, but not class-discriminative. In other words, they are independent of class information. The explanations for different target classes, even randomly chosen classes, are almost identical.

Based on LRP, our work~\cite{Gu2018UnderstandingID} proposes Contrastive Layer-wise Relevance Propagation (CLRP), which is capable of producing instance-specific, class-discriminative, pixel-wise explanations. Before introducing our CLRP, we first discuss the conservative property in the LRP. In a DNN, given the input $\boldsymbol{X} = \{x_1, x_2, x_3,\cdots, x_n\}$, the output $\boldsymbol{Y} = \{y_1, y_2, y_3,\cdots, y_m\}$, the score $S_{y_j}$ (activation value) of the neuron $y_j$ before softmax layer, the LRP generate an explanation for the class $y_j$ by redistributing the score $S_{y_j}$ layer-wise back to the input space. The assigned relevance values of the input neurons are $\boldsymbol{R} = \{r_1, r_2, r_3,\cdots, r_n\}$. The conservative property is defined as follows: The generated saliency map is conservative if the sum of assigned relevance values of the input is equal to the score of the class-specific neuron, $\sum_{i=1}^{n} r_i=S_{y_j}$.

\begin{figure}[t]
    \centering
        \centering
        \includegraphics[scale=0.25]{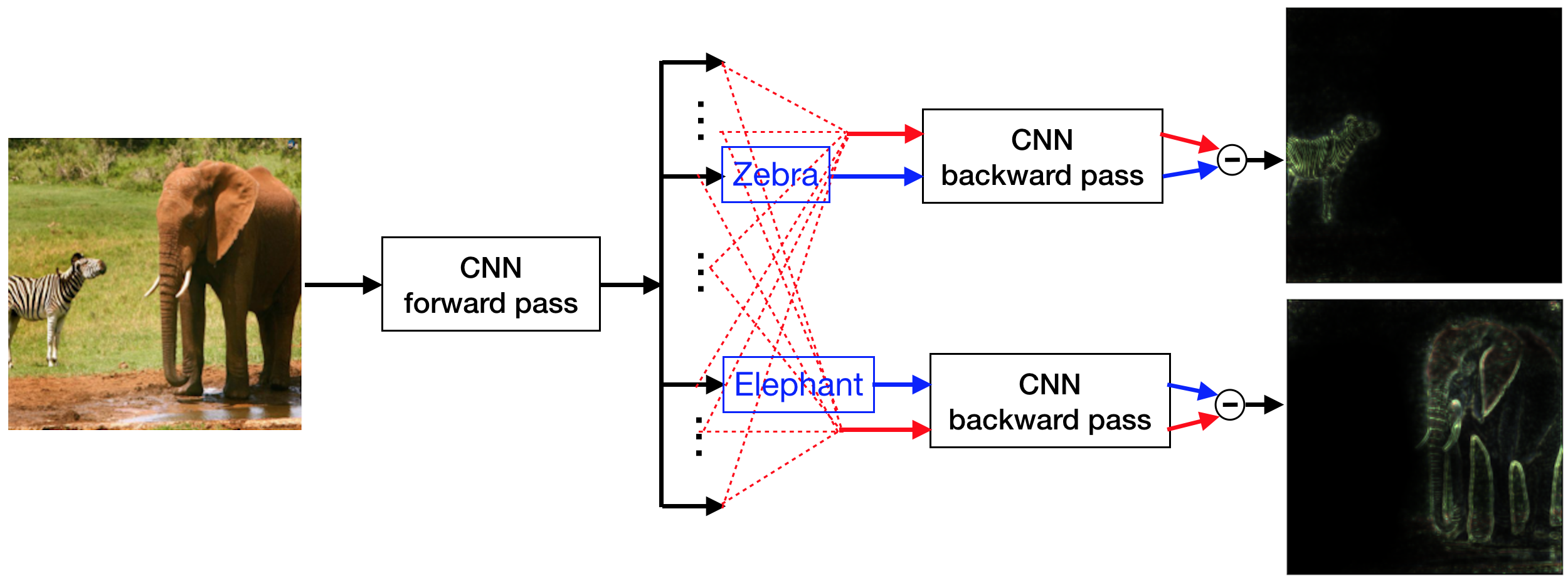}
    \caption{The figure shows an overview of our CLRP. For each predicted class, the approach generates a class-discriminative explanation by comparing two signals. The blue line means the signal that the predicted class represents. The red line models a dual concept opposite to the predicted class. The final explanation is the difference between the two saliency maps that the two signal generate.}
\label{fig:overview_clrp}
\end{figure}

The overview of the CLRP are shown in Fig. \ref{fig:overview_clrp}. We first describe the LRP as follows. The $j$-th class-specific neuron $y_{j}$  is connected to input variables by the weights $\boldsymbol{W}$ of layers between them. The neuron $y_{j}$ models a visual concept $O$. For an input example $\boldsymbol{X}$, the LRP maps the score $S_{y_{j}}$ of the neuron back into the input space to get relevance vector $\boldsymbol{R} = f_{LRP} (\boldsymbol{X}, \boldsymbol{W}, S_{y_{j}})$. In our contrastive LRP, we construct a dual virtual concept $\overline{O}$, which models the opposite visual concept to the concept $O$. For instance, the concept $O$ models the \textbf{zebra}, and the constructed dual concept $\overline{O}$ models the \textbf{non-zebra}. One way to model the $\overline{O}$ is to select all classes except for the target class representing $O$, i.e. the dashed red lines in Fig. \ref{fig:overview_clrp} are connected to all classes except for the target class \textbf{zebra}. Next, the score $S_{y_{j}}$ of target class is uniformly redistributted to other classes. Given the same input example $\boldsymbol{X}$, the LRP generates an explanation $\boldsymbol{R}_{dual} = f_{LRP}(\boldsymbol{X}, \overline{\boldsymbol{W}}, S_{y_{j}})$ for the dual concept. The Contrastive LRP is defined as follows:
\begin{equation}
\boldsymbol{R}_{CLRP} =\max(\boldsymbol{0}, (\boldsymbol{R} - \boldsymbol{R}_{dual}))
\end{equation}
where the function $\max(\boldsymbol{0}, \boldsymbol{X})$ means replacing the negative elements of $\boldsymbol{X}$ with zeros. The difference between the two saliency maps cancels the common parts. Without the dominant common parts, the non-zero elements in $\boldsymbol{R}_{CLRP}$ are the most relevant pixels.

Besides the qualitative evaluation, we also evaluate the explanations quantitatively with a Pointing Game and an ablation study. Both qualitative and quantitative evaluations show that the CLRP generates better explanations than the LRP.

\begin{figure*}[t]
  \centering
   \includegraphics[width=0.85\linewidth]{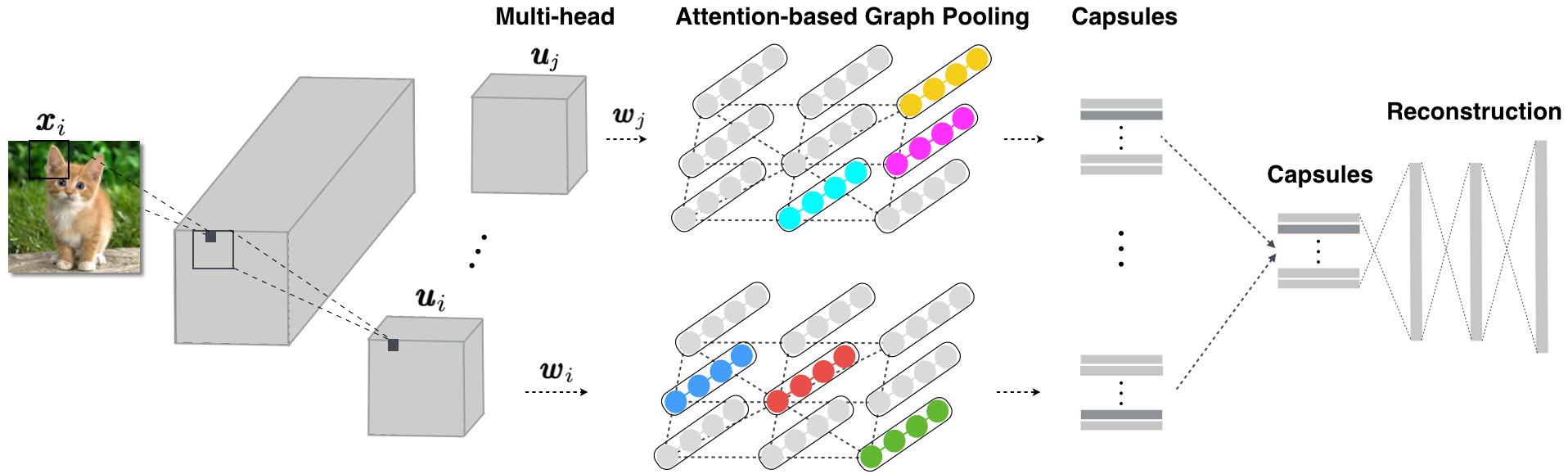}
  \caption{The illustration of GraCapsNets: The extracted primary capsules are transformed and modeled as multiple graphs.  The pooling result on each graph (head) corresponds to one vote. The votes on multiple graphs (heads) are averaged to generate the final prediction.}
  \label{fig:overview_gracaps}
\end{figure*}

\subsection{Explainability of Capsule Network-based Classification}
\label{subsec:caps_exp}
Capsule Networks, as alternatives to Convolutional Neural Networks, have been proposed to recognize objects from images. The current literature demonstrates many advantages of CapsNets over CNNs. However, how to create explanations for individual classifications of CapsNets has not been well explored. 

The widely used saliency methods are mainly proposed for explaining CNN-based classifications; they create saliency map explanations by combining activation values and the corresponding gradients, e.g., Grad-CAM. They combine activation values and the received gradients in specific layers, e.g., deep convolutional layers. In CapsNets, instead of deep convolutional layers, an iterative routing mechanism is applied to extract high-level visual concepts. Hence, these saliency methods cannot be trivially applied to CapsNets. Besides, the routing mechanism makes it more challenging to identify interpretable input features relevant to a classification.

To overcome the lack of interpretability, we can either propose new post-hoc interpretation methods for CapsNets or modify the model to have build-in explanations. In our published work~\cite{gu2020interpretable}, we explore the latter. Specifically, we propose interpretable Graph Capsule Networks (GraCapsNets), where we replace the routing part with a multi-head attention-based Graph Pooling approach. Our GraCapsNet includes an attention-based pooling module, with which individual classification explanations can be created effectively and efficiently.

As introduced in Background Section, CapsNets start with convolutional layers that convert the input pixel intensities $\mathbf{X}$ into primary capsules $\mathbf{u}_i$ (i.e., low-level visual entities). Each $\mathbf{u}_i$ is transformed to vote for high-level capsules $\mathbf{\hat{u}}_{j|i}$ with learned transformation matrices. Then, a routing process is used to identify the coupling coefficients $c_{ij}$, which describe how to weight votes from primary capsules. Finally, a squashing function is applied to the identified high-level capsules $\mathbf{s}_j$ so that the lengths of them correspond to the confidence of the class's existence.
 
Different routing mechanisms differ only in how to identify $c_{ij}$. Routing processes describe one way to aggregate information from primary capsules into high-level ones. In our GraCapsNets, we implement the information aggregation by multi-head graph pooling processes. In CapsNets, the primary capsules represent object parts, e.g., the eyes and nose of a cat. In our GraCapsNets, we explicitly model the relationship between the primary capsules (i.e., part-part relationship) with graphs. Then, the followed graph pooling operations pool relevant object parts from the graphs to make a classification vote. Since the graph pooling operation reveals which input features are pooled as relevant ones, we can easily create explanations to explain the classification decisions.

The overview of our GraCapsNets is illustrated in Fig. \ref{fig:overview_gracaps}. In GraCapsNet, the primary capsules $\mathbf{u}_i$ are transformed into a feature space. All transformed capsules $\mathbf{u}^\prime_{i}$ are modeled as multiple graphs. Each graph corresponds to one head, the pooling result on which corresponds to one vote. The votes on multiple heads are averaged as the final prediction.

The transformed capsules $\mathbf{u}^\prime_{i}$ can be modeled as multiple graphs. A graph consists of a set of nodes and a set of edges. As shown in Fig. \ref{fig:overview_gracaps}, the primary capsules are reshaped from $L$ groups of feature maps. Each group consists of $C$ feature maps of the size $K\times K$. Correspondingly, the transformed capsules $\mathbf{u}^\prime_{i}$ where $i \in \{ 1, 2, ... K^2\}$ form a single graph with $K^2$ nodes. Each node corresponds to one transformed capsule $\mathbf{u}^\prime_{i}$, and the activation vector of $\mathbf{u}^\prime_{i}$ is taken as features of the corresponding node. The graph edge can be represented by an adjacency matrix, where different priors can be modeled. The spatial relationship between primary capsules is modeled in our work.

Given node features $\mathbf{X}^l \in \mathbb{R}^{(K^2 \times D_{out})}$ and adjacency matrix $\mathbf{A}\in \mathbb{R}^{(K^2 \times K^2)}$ in the $l$-th head of GraCapsNet. We first compute the attention of the head as $\mathbf{Att}^l = \mathrm{softmax}(\mathbf{A} \mathbf{X}^l \mathbf{W})$ where $\mathbf{W} \in\mathbb{R}^{D_{out} \times M} $ are learnable parameters. $D_{out}$ is the dimension of the node features and $M$ is the number of output classes. The output is of the shape $(K^2 \times M)$. In our GraCapsNet for object recognition, $\mathbf{Att}^l$ corresponds to the visual attention of the heads.
The graph pooling output $\mathbf{S}^l \in \mathbb{R}^{(M \times D_{out})}$ of the head is computed as $\mathbf{S}^l = (\mathbf{Att}^l)^T \mathbf{X}^l$.
The final predictions of GraCapsNets are based on all $L$ heads with outputs $\mathbf{S}^l$ where $l \in \{1, 2, ..., L\}$. The output capsules are $\mathbf{V} = \mathrm{squash}(\frac{1}{L} \sum^L_{l=1} \mathbf{S}^l)$.

\begin{figure*}[t]
  \centering
   \includegraphics[width=0.6\linewidth]{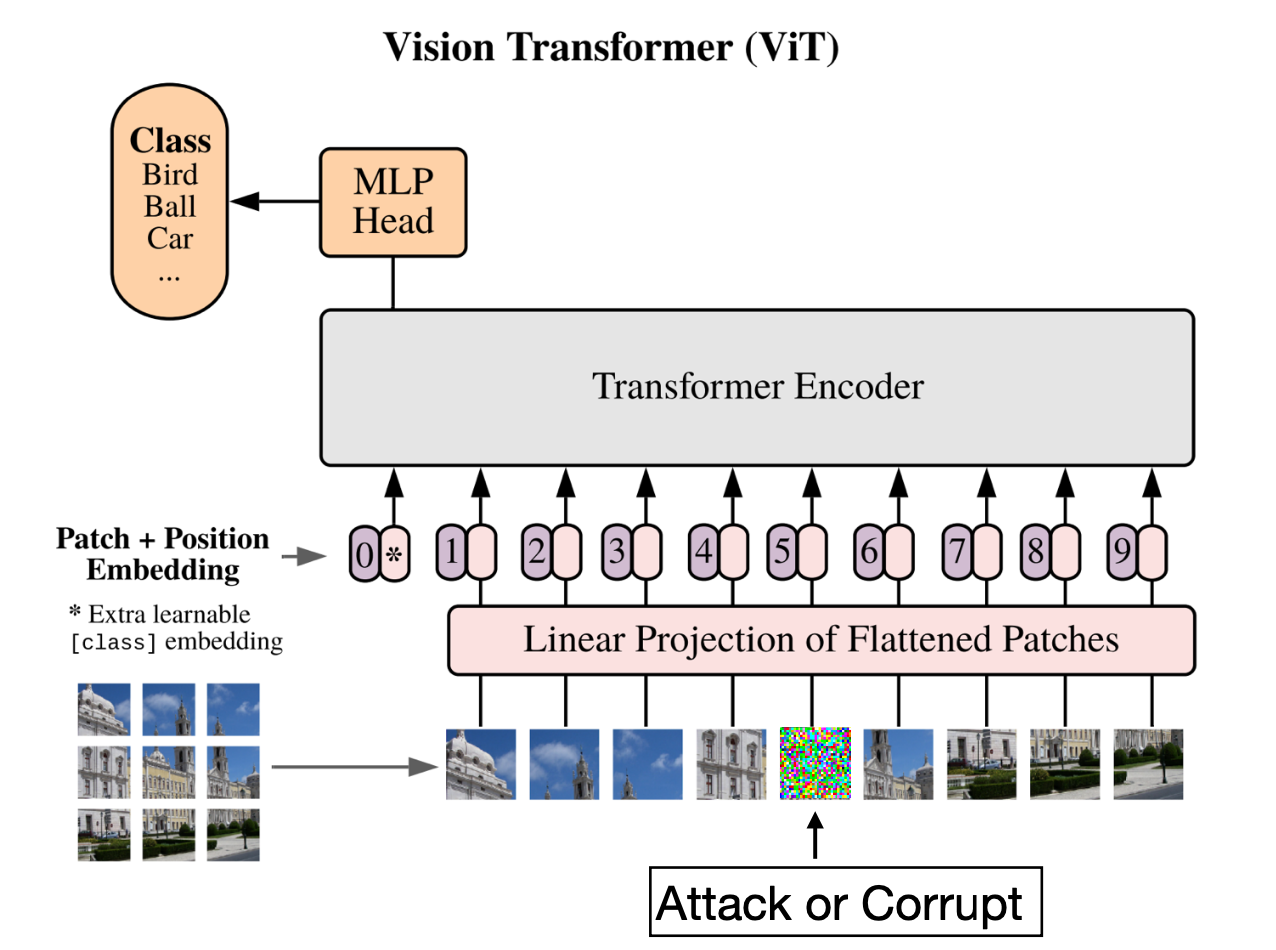}
  \caption{Adversarial Patch Attack or Natural Patch Corruption on Vision Transformer.}
  \label{fig:overview_vit_attack}
\end{figure*}

In our GraCapsNet, we can use visual attention as built-in explanation to explain the predictions of GraCapsNets. The averaged attenion over $l$ heads is
\begin{equation}
\mathbf{E} = \frac{1}{L} \sum^L_{l=1} \mathbf{Att}^l
\end{equation}
where $\mathbf{Att}^l$ corresponds to the attention of the $l$-th head. The created explanations $\mathbf{E}$ are of the shape $(K^2 \times M)$. Given the predicted class, the $K\times K$ attention map indicates which pixels of the input image support the prediction.

The explanations for individual classifications of GraCapsNets can be created in an effective and efficient way. Surprisingly, without a routing mechanism, our GraCapsNets can achieve better classification performance and better adversarial robustness, and still keep other advantages of CapsNets, namely, disentangled representations and affine transformation robustness.

\subsection{Explainability of Vision Transformer-based Classification}
\label{subsec:vit_exp}
The recent advances in Vision Transformer (ViT) have demonstrated its impressive performance in image classification~\cite{dosovitskiy2020image,touvron2021training}, which makes it a promising alternative to Convolutional Neural Network (CNN). Unlike CNNs, ViT represents an input image as a sequence of image patches. Then, a self-attention mechanism is applied to aggregate information from all patches. The attention can be used to create saliency maps to explain ViT-based classification decisions, e.g. with Rollout Attention method~\cite{abnar2020quantifying}. The patch-wise input image representation in ViT makes the following question interesting: How does the attention of ViT change when individual input image patches are perturbed with natural corruptions or adversarial perturbations? For example, Fig. \ref{fig:overview_vit_attack} illustrates the case where a single patch of the input is perturbed or attacked.

\begin{figure}[t]
    \hspace{0.1cm}
    \begin{subfigure}[b]{0.17\textwidth}
    \centering
    \includegraphics[scale=0.25]{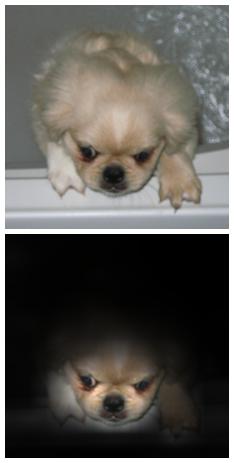}
    \caption{Clean Image}
    \label{fig:teaser_clean}
    \end{subfigure}
    \begin{subfigure}[b]{0.41\textwidth}
    \centering
    \includegraphics[scale=0.25]{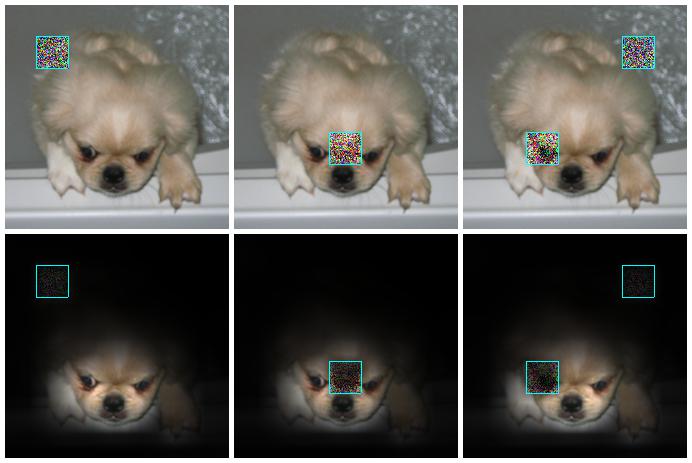}
    \caption{with Naturally Corrupted Patch}
    \label{fig:teaser_nat}
    \end{subfigure}
    \begin{subfigure}[b]{0.41\textwidth}
    \centering
    \includegraphics[scale=0.25]{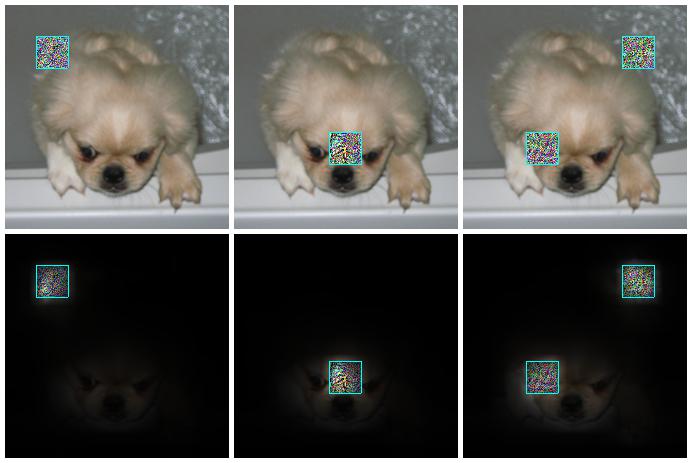}
    \caption{with Adversarial Patch}
    \label{fig:teaser_adv}
    \end{subfigure}
    \caption{Images with patch-wise perturbations (top) and their corresponding attention maps (bottom). The attention mechanism in ViT can effectively ignore the naturally corrupted patches to maintain a correct prediction, whereas it is forced to focus on the adversarial patches to make a mistake. The images with corrupted patches are all correctly classified. The images with adversary patches in subfigure~\ref{fig:teaser_adv} are misclassified as \textit{dragonfly}, \textit{axolotl}, and \textit{lampshade}, respectively.}
    \label{fig:teaser}
\end{figure}

In our work~\cite{gu2021vision}, we study the robustness of vision transformers to patch-wise perturbations. Surprisingly, we find that vision transformers are more robust to naturally corrupted patches than CNNs, whereas they are more vulnerable to adversarial patches. Furthermore, we conduct extensive qualitative and quantitative experiments to understand the classification under patch perturbations. 

We have revealed that ViT's stronger robustness to natural corrupted patches and higher vulnerability against adversarial patches are both caused by the attention mechanism. Specifically, the attention model can help improve the robustness of vision transformers by effectively ignoring natural corrupted patches. However, when vision transformers are attacked by an adversary, the attention mechanism can be easily fooled to focus more on the adversarially perturbed patches and cause a mistake.

Digging down further, we find the reason behind this is that the self-attention mechanism of ViT can effectively ignore the natural patch corruption, while it's also easy to manipulate the self-attention mechanism to focus on an adversarial patch. This is well supported by rollout attention visualization~\cite{abnar2020quantifying} on ViT. As shown in Fig.~\ref{fig:teaser} (a), ViT successfully attends to the class-relevant features on the clean image, i.e., the head of the dog. When one or more patches are perturbed with natural corruptions, shown in Fig.~\ref{fig:teaser} (b), ViT can effectively ignore the corrupted patches and still focus on the main foreground to make a correct prediction. In Fig.~\ref{fig:teaser} (b), the attention weights on the positions of naturally corrupted patches are much smaller even when the patches appear in the foreground. In contrast, when the patches are perturbed with adversarial perturbations by an adversary, shown in Fig.~\ref{fig:teaser} (c), ViT is successfully fooled to make a wrong prediction because the attention of ViT is misled to focus on the adversarial patches instead.

In our work~\cite{gu2021vision}, we provide our understanding of the attention changes of ViT when individual input image patches are perturbed with natural corruptions or adversarial perturbations.

  \section{Robustness of Deep Visual Classification Models}
\label{sec:robust}

\subsection{Introduction}

\begin{table}[]
\footnotesize
\centering
\begin{tabular}{ | c | c | c | m{7.2cm} | }
\hline
\multirow{2}{*}{\thead{Natural \\ Robustness}} & Additive & Natural Corruption & Robustness to the noisy images that are added with various noise~\cite{lee2020compounding,hendrycks2019benchmarking}, such as, white noise, blur, weather, and digital categories. \\
\cline{2-4} 
& Non-Additive & Affine Transformation & Robustness to the images that are affine-transformed from standard ones~\cite{chang2018broadcasting,sabour2017dynamic,gu2020improving}. \\
\hline 
 & \multirow{6}{*}{Additive} & Dense Attack & Robustness to the images where all pixels can be changed under a certain constraint~\cite{goodfellow2001explaining,madry2017towards}. \\
\cline{3-4} 
\multirow{5}{*}{\thead{Adversarial \\ Robustness}} &  & Sparse Attack & Robustness to the images where only a few pixels of each image can be manipulated~\cite{papernot2016limitations}. \\
\cline{3-4} 
&  & Patch Attack &  Robustness to the perturbed images where only a single patch (a specific region) of each image can be manipulated~\cite{brown2017adversarial,karmon2018lavan}. \\
\cline{2-4}
& \multirow{3}{*}{Non-Additive} & \thead{Transformation \\ -Based Attack} & Robustness to adversarial images that is created by delicated affine transformations~\cite{xiao2018spatially}. \\
\cline{3-4} 
&  & Sementic Attack & Robustness to semantic adversarial images that is created by image synthesis~\cite{hosseini2018semantic}. \\
\hline
\end{tabular}
\caption{Categorization of Robustness in Image Classification Task.} 
\label{tab:summary_robust}
\end{table}

In this thesis, we mainly consider two types of robustness, namely, natural robustness and adversarial robustness. When an image is captured, different corruption can happen, e.g., the existence of white noise, the effect of weather, the compression in the digitalization process, and random affine transformation. The robustness to these images with natural corruption is denoted as natural robust. Adversarial robustness describes the robustness of models to adversarial images, which is created by an adversary. Both natural robustness and adversarial robustness are critical in some safety-critical domains. We summarize and categorize the robustness in Tab.~\ref{tab:summary_robust}.

Besides the type of attacks in Tab.~\ref{tab:summary_robust}, adversarial attacks can be categorized into targeted and untargeted ones. The goal of targeted attacks is to mislead the model to a specific target class, while the goal of untargeted ones is to fool the model to make wrong predictions.

In terms of the availability of the target models, adversarial attacks can also be categorized into white-box and black-box attacks. The white-box attacks assume that the adversary has all access to target models including model parameters, model architectures, and even defense methods. In contrast, in the setting of black-box attacks, the adversary can only obtain the output of the target model. The black-box attacks have also received great attention since it is realistic in real-world applications.

The implementation of white-box attacks is relatively cheap where they create adversarial examples with the gradients of the self-defined objective function with respect to inputs. However, the implementation of black-box attacks can be computationally expensive given the limited available information. One way to created adversarial examples in a black-box fashion is to leverage their transferability~\cite{liu2016delving,xie2019improving,dong2019evading,zou2020improving,guo2020backpropagating,wu2020skip,huang2019enhancing,inkawhich2020perturbing,li2020learning,wang2020unified,gu2021adversarial}, namely, the adversarial examples created on one model can also fool another. The adversary first trains a surrogate model on the same training data as the one used for the target model and creates adversarial examples on the surrogate model to fool the target model, which is called transfer-based black-box attack. However, the transfer-based black-box attacks require access to the training data of the target model. To overcome the limitation, the query-based black-box attacks have been proposed where the attacks are based on the outputs obtained by querying the target models directly~\cite{chen2017zoo,cheng2018query,bhagoji2018practical,cheng2018query,andriushchenko2020square}.

In addition, based on the constraints on the adversarial images, the generated adversarial perturbations can be quasi-imperceptible or unbounded. The popular metric of to measure the distance between clean images and adversarial image is $\ell_p$ norm~\cite{papernot2016limitations}, such as, $\ell_{1}$, $\ell_{2}$ and $\ell_{\infty}$. However, the metric is not perfectly aligned with human perception. The more advanced metric has also been explored in the community, e.g., Wasserstein distance~\cite{wong2019wasserstein}.

Given the potential threats posed by adversarial attacks, many defense strategies have been proposed to build adversarially robust models. One of the most effective defense methods is adversarial training, which creates adversarial examples and adds them to the training dataset in each training iteration. Besides, the pre-processing methods have been explored to purify adversarial examples~\cite{goodfellow2014explaining,tramer2017ensemble,carlini2017towards,shafahi2019adversarial,zhang2019you,zhang2019theoretically,andriushchenko2020understanding,kim2020understanding,wong2020fast,rice2020overfitting,vivek2020single,lee2020adversarial,wang2021convergence,sriramanan2021towards,wu2021attacking}. However, some of the defense strategies have broken again in later publications~\cite{athalye2018obfuscated}. Some defense methods provide certified robustness to break arm-race between adversary and defense~\cite{raghunathan2018certified,jia2019certified,cohen2019certified,li2019certified,salman2019convex,salman2019provably,mohapatra2020higher,salman2020denoised,gu2022segpgd,wu2022towards}. Even many methods have been published to address, the accuracy of the model under attacks is still much lower than the accuracy on clean images, especially on the large dataset~\cite{xie2019feature}. In addition to building robust model, another way to address the threats is to detect adversarial examples first~\cite{xu2017feature,feinman2017detecting,pang2018towards,lee2018simple,zheng2018robust,roth2019odds,cohen2020detecting}.

%\textbf{Contributions.} In this subsection, we categorize the robustness of image classifications. In this thesis, our contributions mainly focus on the role the model architecture plays in terms of both natural robustness and adversarial robustness. Specifically, we introduce and summarize our studies on the robustness of Capsule Networks in subsection \ref{subsec:caps_robust}. We elaborate our findings and improvements of Capsule Networks' robustness to non-additive perturbation in Chapter 4 and 5, and our Capsule Network attack method, called Vote Attack, in Chapter 6. Besides, we introduce our understanding of the robustness of ViT-based classifications to patach-wise perturbations in subsection \ref{subsec:caps_robust}, which is elaborated in Chapter 7.

In this subsection, we categorize the robustness of image classifications. Our contributions of this thesis mainly focus on the role the model architecture plays in terms of both natural robustness and adversarial robustness. In the rest of this section, we present our contributions towards the robustness of image classification models, such as Capsule Networks and Vision Transformers.

\subsection{Robustness of Capsule Network-based Classification}
\label{subsec:caps_robust}
Human visual recognition is quite insensitive to affine transformations. For example, entities in an image, and a rotated version of the entities in the image, can both be recognized by the human visual system, as long as the rotation is not too large. Convolutional Neural Networks (CNNs), the currently leading approach to image analysis, achieve affine robustness by training on a large amount of data that contain different transformations of target objects. Given limited training data, a common issue in many real-world tasks, the robustness of CNNs to novel affine transformations is limited~\cite{sabour2017dynamic}.

With the goal of learning image features that are more aligned with human perception, Capsule Networks (CapsNets) have recently been proposed~\cite{sabour2017dynamic}. Our work~\cite{gu2020improving} first investigates the effectiveness of components that make CapsNets robust to input affine transformations, with a focus on the routing algorithm. However, recent work~\cite{paik2019capsule} shows that all routing algorithms proposed so far perform even worse than a uniform/random routing procedure.

From both numerical analysis and empirical experiments, our investigation reveals that the dynamic routing procedure contributes neither to the generalization ability nor to the affine robustness of CapsNets. Therefore, it is infeasible to improve the affine robustness by modifying the routing procedure. Instead, we investigate the limitations of the CapsNet architectures and propose a simple solution. Namely, we propose to apply an identical transformation function for all primary capsules and replace the routing with a simple averaging procedure.

Besides the high affine transformation robustness, CapsNets also demonstrate other advantages, such as the ability to recognize overlapping digits and the semantic representation compactness. In recent years, It has been suggested that CapsNets have the potential to surpass the dominant convolutional networks in these aspects~\cite{sabour2017dynamic,hinton2018matrix,rajasegaran2019deepcaps,gu2021effective}. However, there lack of comprehensive comparisons to support this assumption, and even for some reported improvements, there are no solid ablation studies to figure out which ones of the components in CapsNets are, in fact, effective.

In our work~\cite{gu2021capsule}, we first carefully examine the major differences in design between the capsule networks and the common convolutional networks adopted for image classification. The difference can be summarized as \textit{a non-shared transformation module, a dynamic routing layer to automatically group input capsules to produce output capsules, a squashing function, a marginal classification loss, and a class-conditional reconstruction sub-network with a reconstruction loss}.

Unlike previous studies~\cite{sabour2017dynamic,hinton2018matrix} which usually take CapsNet as a whole to test its robustness, our work~\cite{gu2021capsule} instead tries to study the effects of each of the above components in their effectiveness on robustness. We consider the three different aspects, such as the robustness to affine transformations, the ability to recognize overlapping digits,
and the semantic representation compactness.

Our investigations reveal that some widely believed benefits of Capsule networks could be wrong:
\begin{enumerate}
    \item The dynamic routing actually may harm the robustness to input affine transformation, in contrast to the common belief;
    \item The high performance of CapsNets to recognize overlapping digits can be mainly attributed to the extra modeling capacity brought by the transformation matrices.
    \item Some components of CapsNets are indeed beneficial for learning semantic representations, e.g., the conditional reconstruction and the squashing function, but they are mainly auxiliary components and can be applied beyond CapsNets.
\end{enumerate}

In addition to these findings, we also enhance common ConvNets by the useful components of CapsNet, and achieve greater robustness. Our investigation shows that Capsule Network is not more robust than Convolutional Network.

\subsection{Robustness of Vision Transformer-based Classification}
\label{subsec:vit_robust}
CapsNets with brain-inspired architectures have more inductive bias than CNNs. Different from CapsNet, Vision Transformer (ViT)  \cite{dosovitskiy2020image} has less architecture bias than CNNs. ViT processes the input image as a sequence of image patches. Then, a self-attention mechanism is applied to aggregate information from all patches. Existing works have shown that ViTs are more robust than CNNs when the whole input image is perturbed with natural corruptions or adversarial perturbations~\cite{bhojanapalli2021understanding,shao2021adversarial,paul2021vision}. Given the patch-based architecture of ViT, our work studies the robustness of ViT to patch-based perturbation.

Two typical types of perturbations are considered to compare the robustness between ViTs and CNN (e.g., ResNets~\cite{he2016deep}). One is natural corruptions~\cite{hendrycks2019benchmarking}, which is to test models' robustness under distributional shift. The other is adversarial perturbations~\cite{szegedy2013intriguing,goodfellow2014explaining}, which are created by an adversary to specifically fool a model to make a wrong prediction. 

We reveal that ViT does not always perform more robustly than ResNet. When individual image patches are naturally corrupted, ViT performs more robustly than ResNet. However, when input image patch(s) are adversarially attacked, ViT shows a higher vulnerability. Digging down further, we find the reason behind this is that the self-attention mechanism of ViT can effectively ignore the natural patch corruption, while it's also easy to manipulate the self-attention mechanism to focus on an adversarial patch. 

Based on the patch-based architectural structure of vision transformers, we further investigate the sensitivity of ViT against patch positions and patch alignment of adversarial patches. First, we discover that ViT is insensitive to different patch positions, while ResNet shows high vulnerability on the central area of input images and much less on corners. We attribute this to the architecture bias of ResNet where pixels in the center can affect more neurons than the ones in corners. In contrast, each patch within ViT can equally interact with other patches regardless of its position.
Further, we find that for ViT, the adversarial perturbation designed to attack one particular position can successfully transfer to other positions of the same image as long as they are aligned with input patches. In contrast, the ones on ResNet hardly do.

To summarise, in our work~\cite{gu2021vision}, we compare ViT and CNNs in terms of the robustness to natural patch corruptions or adversarial patch attacks.
  
\bibliographystyle{unsrt}
\bibliography{literatur}

\end{document}